\documentclass[11pt,a4paper]{article}
\usepackage[hyperref]{acl2017}
\usepackage{times}
\usepackage{latexsym}

\usepackage{url}
\usepackage{graphicx}

\aclfinalcopy 

\title{NILC-USP at SemEval-2017 Task 4: A Multi-view Ensemble for Twitter Sentiment Analysis}

\author{Edilson A. Corr\^ea Jr., Vanessa Queiroz Marinho, Leandro Borges dos Santos\\
  Institute of Mathematics and Computer Science \\
  University of S\~ao Paulo (USP)\\
  S\~ao Carlos, S\~ao Paulo, Brazil\\
  {\tt \{edilsonacjr,vanessaqm,leandrobs\}@usp.br} \\}

\date{}

\begin{document}
\maketitle
\begin{abstract}
  This paper describes our multi-view ensemble approach to SemEval-2017 Task 4 on Sentiment Analysis in Twitter, specifically, the Message Polarity Classification subtask for English (subtask A). 
Our system is a voting ensemble, where each base classifier is trained in a different feature space. The first space is a bag-of-words model and has a Linear SVM as base classifier. The second and third spaces are two different strategies of combining word embeddings to represent sentences and use a Linear SVM and a Logistic Regressor as base classifiers. 
The proposed system was ranked 18th out of 38 systems considering F1 score and  20th considering recall.


\end{abstract}

\section{Introduction}
Twitter is a microblogging service that has 313 million monthly active users~\footnote{https://about.twitter.com/company}. In this social media platform, users interact through short messages, so-called \emph{tweets}.
The company estimates that over 500 million tweets are sent each day~\footnote{https://business.twitter.com/en/basics.html}. Despite of their size, at most 140 characters, these messages provide rich data because users generally write about their thoughts, opinions and sentiments. 
Therefore, applications in several domains, such as commercial~\cite{Jansen:2009:Twitter} and political~\cite{tumasjan2010predicting,Wang2012system}, may benefit from the automatic classification of sentiment in tweets. 

In this paper we show that a multi-view ensemble approach that leverages simple representations of texts may achieve good results in the task of message polarity classification. 
The proposed system consists of three base classifiers, each of them with a specific text representation technique (bag-of-words or word embeddings combination). As base classifiers, we use Support Vector Machines (SVM) and Logistic Regression.
The proposed approach was evaluated on the SemEval-2017 Task 4 Subtask A for English and was ranked 18th out of 38 participating systems considering F1 score and 20th considering recall.

This paper is organized as follows: Section 2 describes our system, the feature spaces and the classifiers we employed. The training and evaluation datasets are presented in Section 3. In addition, Section 3 also describes the preprocessing steps and some details about the word embeddings. We present our results in Section 4. Finally, Section 5 outlines our conclusions and remarks on future work.
\section{System Description}

The proposed system consists of a multi-view ensemble with three base classifiers with different text representation techniques (feature spaces), that is, all base classifiers are trained on the same dataset but with a different representation or feature space. The first is a Linear SVM and the tweets are represented using bag-of-words (Section \ref{sec:bag}). The second is another Linear SVM and the tweets are represented by averaging the word embeddings (Section \ref{sec:emb}). The third is a Logistic Regressor and the tweets are represented by averaging the weighted word embeddings (Section \ref{sec:emb}). In these systems, a class of a given instance is decided as the class which maximizes the sums of the predicted probabilities (soft-voting).


The idea of using a multi-view ensemble is to explore different feature spaces without the need of combining all features in the same space, since this combination may lead to the insertion of noise. Moreover, there is no straightforward way of combining them. Other important aspect of this technique is the possibility of assigning different weights to each pair of classifier/features (view) or even learn the weights by a regression method~\cite{xu2013multiview}.

The process of choosing the best classifier for each feature space was done by executing a Grid Search with state-of-art classifiers ($k$-NN, Decision Trees, Naive Bayes, Random Forest, AdaBoost, SVM and Logistic Regression), which led to the selection of SVM and Logistic Regression. In the following subsections, the text representations and the classifiers are described. The complete system will be made available~\footnote{https://github.com/edilsonacjr/semeval2017}.


\subsection{Text representations}

In this work, tweets were modeled using three types of text representation. The first one is a bag-of-words model weighted by \textit{tf-idf} (term frequency - inverse document frequency) (Section \ref{sec:bag}). The second represents a sentence by averaging the word embeddings of all words (in the sentence) and the third represents a sentence by averaging the weighted word embeddings of all words, the weight of a word is given by \textit{tf-idf} (Section \ref{sec:emb}). 


\subsubsection{Bag-of-words}\label{sec:bag}

The bag-of-words model is a popular approach to represent text (documents, sentences, queries and others) in Natural Language Processing and Information Retrieval. In this model, a text is represented by its set of words, this representation can be binary, in which a word receives 1 if it is in the text or 0 otherwise, considering a predefined vocabulary. An alternative is a representation weighted by some specific information, such as frequency. The representation adopted by this work is the bag-of-words weighted by \textit{tf-idf}~\cite{salton1989automatic}, where each tweet is represented as $tweet_i = (a_{i1}, a_{i2},...,a_{im} )$, where $a_{ij}$ is given by the frequency of term $t_{j}$ in the tweet $i$ weighted by the total number of tweets divided by the amount of tweets containing the term $t_{j}$.


\subsubsection{Word embeddings}\label{sec:emb}

Word embeddings, a concept introduced by~\citet{bengio2003neural}, is a distributional representation of words, where each word is represented by a dense, real-valued vector. These vectors are learned by neural networks trained in language modeling~\cite{bengio2003neural} or similar tasks~\cite{collobert2011nlp,mikolov2013efficient,mikolov2013distributed}. In this work, the Word2Vec model~\cite{mikolov2013efficient,mikolov2013distributed} is used, in which the vectors are learned by training the neural network to perform context (skip-gram) or word (CBOW) prediction. 

In addition to capture syntactic and semantic information, the vectors produced by Word2Vec have geometric properties such as \textit{compositionality}~\cite{mikolov2013distributed}, which allow larger blocks of information (such as sentences and paragraphs) to be represented by the combination of the embeddings of the words contained in the block. This approach has been adopted by a considerable number of works in several areas, such as question answering~\cite{belinkov2015vectorslu}, semantic textual similarity~\cite{sultan2015dls}, word sense disambiguation~\cite{iacobacci2016embeddings} and even sentiment analysis~\cite{socher2013recursive}. 

In our work we adopted two combination approaches. The first is a simple combination, where each tweet is represented by the average of the word embedding vectors of the words that compose the tweet. The second approach also averages the word embedding vectors, but each embedding vector is now weighted (multiplied) by the \textit{tf-idf} of the word it represents. A similar approach has been used in~\citet{zhao2015ecnu}.





\subsection{Classifiers}

For both classifiers we used the well-known \textit{Scikit-learn}~\cite{pedregosa2011scikit} implementation (with default parameters).



\subsubsection{Logistic Regression}

Logistic Regression is a linear classifier that predicts the class probabilities of a binary classification problem. It is also known as logit regression because a sigmoid function outputs the class probabilities. To tackle multiclass problems, the training algorithm uses the one-vs-rest approach~\cite{Murphy:2012:Machine:Learning}.

\subsubsection{Support Vector Machines}

Support Vector Machines (SVMs) classifiers find the decision boundary that maximizes the margin between two classes. However, when data is intrinsically nonlinear, SVM classifiers cannot properly separate between classes. A possible solution is to map the data points into a higher-dimensional feature space. By doing so, the data becomes linearly separable \cite{Murphy:2012:Machine:Learning}. To apply the algorithm to our multiclass problem, we used the one-vs-one approach.


\section{Data}

To evaluate our system we used the training and development datasets provided by the SemEval-2017 competition (specifically Twitter2016-train and Twitter2016-dev). For testing, in addition to the previous year's testing datasets (Twitter2016-test, SMS2013, Tw2014-sarcasm, LiveJournal2014), a new dataset (Twitter2017-test) was made available. A summary of the datasets is given in Table \ref{tab:datasests}.

\begin{table}[h]
\begin{center}
\resizebox{0.48\textwidth}{!}{
\begin{tabular}{lllll}
\bf Dataset & \bf Total & \bf Pos. & \bf Neg. & \bf Neut. \\ 
\hline
Twitter2016-train 	& $6000$ & $3094$ & $863$ & $2043$\\
Twitter2016-dev 	& $2000$ & $844$ & $765$ & $391$\\
Twitter2016-test 	& $20632$ & $7059$ & $10342$ & $3231$\\
Twitter2017-test 	& $12284$ & $2375$ & $3972$ & $5937$\\
SMS2013 		& $2093$ & $492$ & $394$ & $1207$\\
Tw2014-sarcasm & $86$ & $33$ & $40$ & $13$\\
LiveJournal2014 & $1142$ & $427$  & $304$ & $411$\\
\hline

\end{tabular}}
\end{center}

\caption{\label{tab:datasests}Datasets used in the training and evaluation of the system.}
\end{table}

\paragraph{Data preprocessing.} Before extracting features, the tweets were preprocessed. First, we tokenized the text considering HTML tags, mentions/usernames, URLs, numbers, words (including hyphenated words) and emoticons. Then, the text was set to lowercase and stopwords (words with low semantic value such as prepositions and articles), punctuation marks, hashtags and mentions/usernames were removed.

\paragraph{Word embeddings.} We used the pre-trained word embeddings~\footnote{code.google.com/archive/p/word2vec/}, generated with the Word2Vec model~\cite{mikolov2013efficient,mikolov2013distributed} and trained on part of Google News dataset, which is composed of approximately 100 billion words. The model comprises 3 million words and phrases and the embedding vectors have 300 dimensions. Words out of the vocabulary were disregarded and when all words in a tweet had no pre-trained vectors, a randomly initialized vector of 300 dimensions was assigned.


\section{Results}

To evaluate, compare and rank the participating systems, F1 score (average), recall (average) and accuracy were chosen by the organizers. Our system was ranked 18th out of 38 systems, with a F1 score of $0.595$ on the Twitter2017-test, ranked 20th considering recall ($0.612$) and ranked 16th with $0.617$ of accuracy. The full ranking and other details of the competition may be found in \citet{SemEval:2017:task4}.

\begin{table}[h]
\begin{center}
\begin{tabular}{lrrr}
\bf Dataset & \bf F1 score & \bf recall & \bf accuracy\\ 
\hline
Twitter2016-test 	& $0.523$ & $0.527$ & $0.542$\\
Twitter2017-test 	& $0.595$ & $0.612$ & $0.617$\\
SMS2013 			& $0.381$ & $0.494$ & $0.609$\\
Tw2014-sarcasm 		& $0.339$ & $0.536$ & $0.442$\\
LiveJournal2014 	& $0.573$ & $0.569$ & $0.588$\\

\hline
\end{tabular}
\end{center}
\caption{\label{tab:scores} Results obtained by our systems in different evaluation datasets.}
\end{table}

In the Twitter2016-test evaluation, only Twitter2016-train and Twitter2016-dev were used in training. In the rest of the evaluations, Twitter2016-train, Twitter2016-dev, Twitter2016-test were used. Despite of the availability of other datasets for training, we chose to use only the three. The results obtained by our system are summarized in Table \ref{tab:scores}.

From the results it is possible to notice that the system is impaired in datasets of different origin, such as SMS2013, this may occur due to the use of a distinct and specific vocabulary. In the case of Tw2014-sarcasm, the major problem is that our representations do not consider the order of words in the sentence which can make it difficult to identify sarcasm or modifiers. In the LiveJournal2014 dataset the system remained stable even though it is a dataset of another domain, probably because it is similar to the Twitter datasets.

\section{Conclusion and Future Work}
In this paper, we presented a multi-view ensemble approach to message polarity classification that participated in the SemEval-2017 Task 4 on Sentiment Analysis in Twitter (subtask A English). Our system was ranked 18th out of 38 participants. 

The results indicated that a multi-view ensemble approach that leverages simple representations of texts may achieve good results in the task of message polarity classification with almost no intervention or special preprocessing. 

In our approach, tweets with opposite polarities might end up with the same vector representation in the cases they present the same words, such as \emph{"It isn't horrible, it's perfect"} and \emph{"It isn't perfect, it's horrible"}. To solve this problem, we plan to combine our model with other techniques that consider the ordering of words, such as word $n$-grams. In the future, we also plan to use approaches for the normalization of informal texts in order to capture particularities of the social media language. In informal texts, a word or a sequence of words can be intentionally replaced, for example \emph{you}, \emph{are} and \emph{see you} can be written as \emph{u}, \emph{r} and \emph{cu}. Because these forms are not mapped into the original words, they are seen as different tokens. In addition, commonly used abbreviations, such as \emph{omg} and \emph{wth}, may express sentiment and their expansion could lead to model improvements. Other improvements that may lead to a better system is the use of word embeddings trained to capture sentiment information and the use of autoencoders to generate sentence/document embeddings.


\section*{Acknowledgments}
E.A.C.J. acknowledge financial support from Google (Google Research Awards in Latin America grant) and CAPES (Coordination for the Improvement of Higher Education Personnel). V.Q.M. acknowledge  financial support from FAPESP  (grant  no. 15/05676-8). L.B.S. acknowledge financial support from Google (Google Research Awards in Latin America grant) and  CNPq (National Council for Scientific and Technological Development, Brazil).

\bibliography{acl2017}
\bibliographystyle{acl_natbib}

\end{document}